\documentclass[conference]{IEEEtran}
\IEEEoverridecommandlockouts
\usepackage{cite}
\usepackage{amsmath,amssymb,amsfonts}
\usepackage{algorithmic}
\usepackage{graphicx}
\usepackage{textcomp}
\usepackage{xcolor}
\def\BibTeX{{\rm B\kern-.05em{\sc i\kern-.025em b}\kern-.08em
    T\kern-.1667em\lower.7ex\hbox{E}\kern-.125emX}}
\begin{document}

\title{Sparsifying and Down-scaling Networks to Increase Robustness to Distortions\\
}

\author{

\IEEEauthorblockN{Sergey Tarasenko}
\IEEEauthorblockA{\textit{System Development Unit, AI System Dept.} \\
\textit{Mobility Technologies}\\
Tokyo, Japan \\
sergey.tarasenko@mo-t.com}
}

\maketitle

\begin{abstract}
It has been shown that perfectly trained networks exhibit drastic reduction in performance when presented with distorted images. Streaming Network (STNet) is a novel architecture capable of robust classification of the distorted images while been trained on undistorted images. The distortion robustness is enabled by means of sparse input and isolated parallel streams with decoupled weights.  Recent results prove STNet is robust to 20 types of noise and distortions.  STNet exhibits state-of-the-art performance for classification of low light images, while being of much smaller size when other networks. In this paper, we construct STNets by using scaled versions (number of filters in each layer is reduced by factor of n) of popular networks like VGG16, ResNet50 and MobileNetV2 as parallel streams. These new STNets are tested on several datasets. Our results indicate that more efficient (less FLOPS), new STNets exhibit higher or equal accuracy in comparison with original networks. Considering a diversity of datasets and networks used for tests, we conclude that a new type of STNets is an efficient tool for robust classification of distorted images.

\end{abstract}

\begin{IEEEkeywords}
streaming networks, STNets, noise robustness, distortion robustenss, state-of-the-art model based STNets
\end{IEEEkeywords}

\section{Introduction}

After introduction of AlexNet \cite{Krizhevsky2012ImageNetCW}, which attracted close attention to the convnets, a vast variety of conv net architectures has been proposed (ResNet\cite{He2015DeepRL}, GoogLeNet \cite{Szegedy2014GoingDW}, VGG16/19 \cite{Simonyan2014VeryDC} etc). Until recently, network performance was improved by means of new model with bigger number of parameters and more sophisticated techniques like skip connections, batch normalization, etc.

Most recent trend in constructing neural networks is to optimize network in term of number of parameters. For example, EfficientNet \cite{Tan2019EfficientNetRM} is optimized in three dimensions, i.e., width (number of filters in convolution layers), depth (number of layers) and input image resolution. Two key takesaway here are 1) it is possible to make more efficient networks with higher accuracy than existing state-of-the-art models; 2) network specs (width and depth) should be considered together with input specs (image resolution). 

On the other hand, conv nets with more than one processing stream have started to gain popularity. To our knowledge, the first two-stream network was introduced by Chorpa \cite{Chopra2005LearningAS} and it is widely known as a ``Siamese network". The motivation behind two streams is that each of the streams carries information about a dedicated image. Images fed to the streams are different. Two-stream networks have been used for the vast variety of tasks (similarity assessment [Siamese networks and pseudo-Siamese \cite{Chopra2005LearningAS, Zagoruyko2015LearningTC}], segmentation-based change detection \cite{Varghese2018ChangeNetAD}, action recognition in videos \cite{Simonyan2014TwoStreamCN}, one-shot image recognition \cite{Koch2015SiameseNN}, simultaneous detection and segmentation \cite{Hariharan2014SimultaneousDA}, human-object interaction recognition \cite{Gkioxari2017DetectingAR}, group activity recognition \cite{Azar2018AMC}, etc. These networks merge different types of information, e.g., image and video, to achieve high accuracy for new tasks.

Another recent trend is testing network performance with adversarial distortions, which are used to check network stability. It was found that even slight distortions can ruin performance of very well tuned networks \cite{Nguyen2015DeepNN,Dodge2017ASA,Hosseini2017GooglesCV}. Therefore effective method to reduce influence of distortions is necessary.

The most recently, a new family of multi-stream network called Streaming Networks (STNets) \cite{Tarasenko2019StreamingNE} has been proposed. It has been shown that cloning the networks into multiple streams and using non-identical intensity slices of the input image as inputs into separate streams increases noise robustness.

In this paper, inspired by optimization behind EffientNet, we propose a practical method to make STNets smaller (number of parameters), faster (FLOPs) and increase noise robustness at the same time.

\section{Prior works}
\label{prior}

It was confirmed that even well trained state-of-the-art models can easily misclassify distorted images \cite{Nguyen2015DeepNN}. Furthermore, several studies have illustrated that performance of conv nets is fragile on various types of distortions, e.g., impulse noise \cite{Hosseini2017GooglesCV}, Gaussian noise and blur \cite{Dodge2017ASA}, various noise types and elastic transforms \cite{Geirhos2017ComparingDN}. There are also numerous studies reporting failure of conv nets due to various types of adversarial attacks \cite{Eykholt2017RobustPA,MoosaviDezfooli2017UniversalAP}.

To enable testing of network robustness to various types of perturbation, Hendrycks and Dietterich \cite{Hendrycks2018BenchmarkingNN} have designed a specialized datasets Cifar10 Corrupted and ImageNet-C/P. Cifar10 Corrupted contains 19 types of distortions: brightness, contrast, defocus blur, elastic transform, fog, frost, gaussian blur, gaussian noise, glass blur, impulse noise, jpeg compression, motion blur, pixelate, saturate, shot noise, snow, spatter, speckle noise, zoom blur.

We refer to brief description of some specific types of noise provided in \cite{Hendrycks2018BenchmarkingNN}(p. 3) : "Shot noise ... is electronic noise caused by the discrete nature of light. Impulse noise is a color analogue of salt-and-pepper noise. Defocus blur occurs when an image is out of focus. Zoom blur occurs when a camera moves toward an object rapidly. Snow is a visually obstructive form of precipitation. Frost forms when lenses or windows are coated with ice crystals. Fog shrouds objects. Brightness varies with daylight intensity. Contrast can be high or low depending on lighting conditions. Elastic transformations stretch or contract small image regions. Pixelation occurs when upsampling a low-resolution image. JPEG is a lossy image compression format that increases image pixelation and introduces artifacts". 

Study \cite{Hendrycks2018BenchmarkingNN} also shares observation that under all equal conditions, simpler models often generalize better and deliver higher robustness to noise.

\begin{figure}
\includegraphics[width=0.9\linewidth]{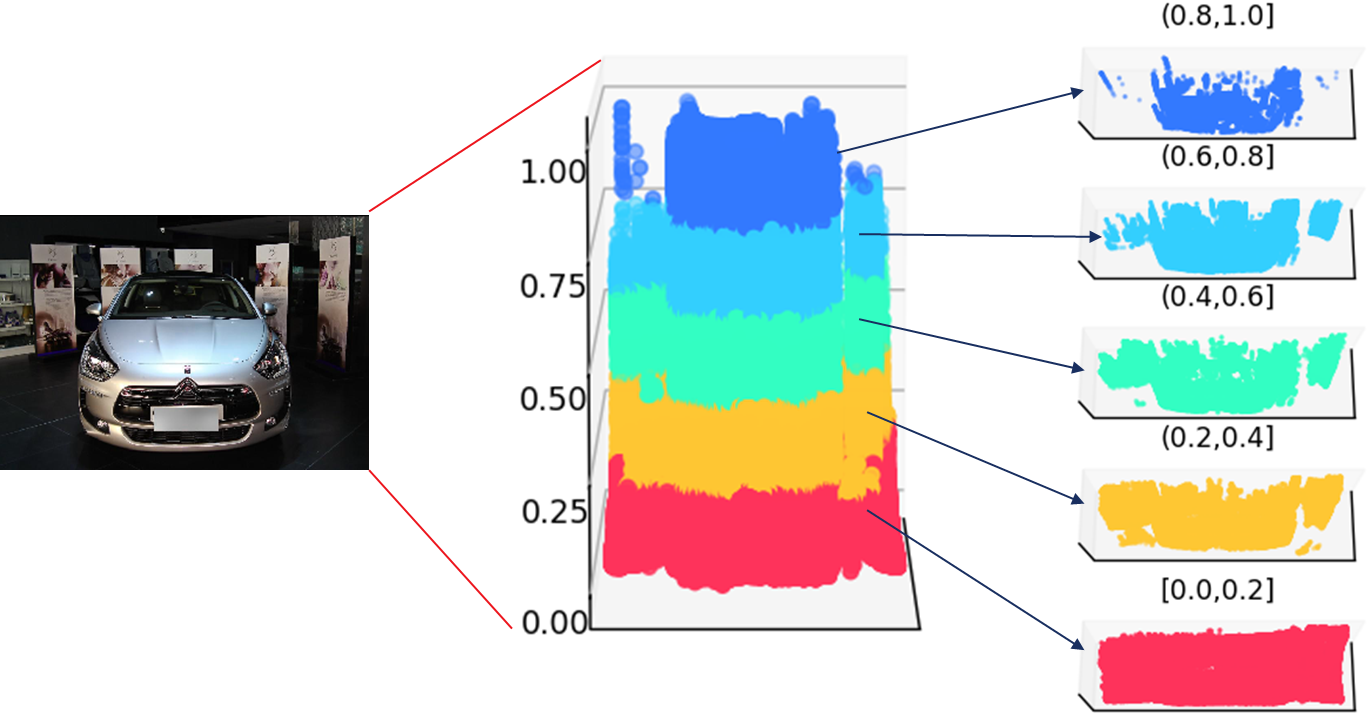}
\caption{Example of intensity slices. }
\label{fig:slices}
\end{figure}

\begin{figure}
\includegraphics[width=1.0\linewidth]{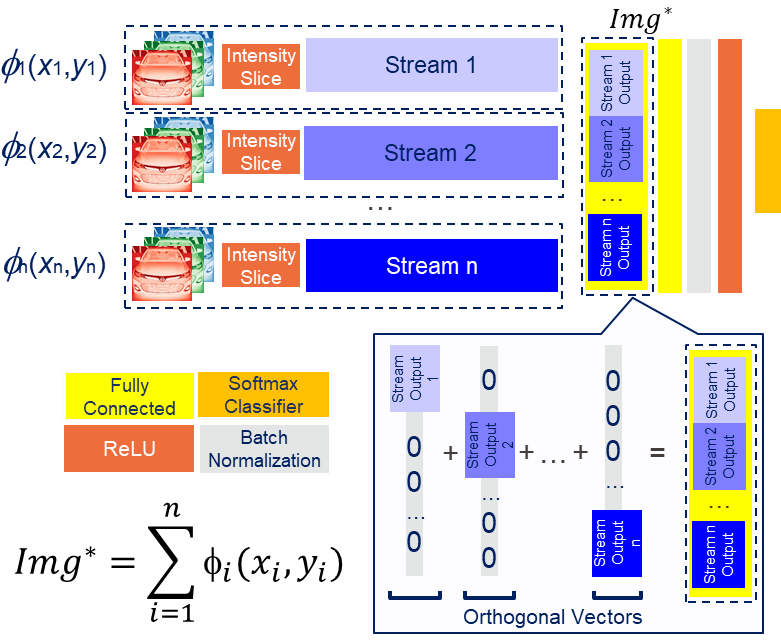}
\caption{STNet architecture brief.}
\label{fig:stnets}
\end{figure}

STNets \cite{Tarasenko2019StreamingNE} has been designed to address the issue of robustness to noise and various types of other distortions. STNet is a neural network, which consist of multiple parallel streams. Each stream has a unique input in a form of an intensity slice of the original input image. Intensity slices are obtained from the original image by keeping pixel values within a certain range and zeroing out others outside the range. Outputs of all streams are concatenated and fed into a single joint classifier. Each stream takes a unique intensity slice of an input image. An example of intensity slices is presented in Fig. \ref{fig:slices} (original image comes from Carvana\footnote{https://www.kaggle.com/c/carvana-image-masking-challenge} dataset). Architecture of the STNet is presented in Fig. \ref{fig:stnets}.

STNet \cite{Tarasenko2019StreamingNE} has confirmed high robustness to noise, when tested on a random zero noise \cite{Tarasenko2020STNets}. A random zero noise implies that values of random pixels are set to 0 for all color channels. STNet's noise robustness has been also confirmed using all 19 types of distortions provided by Cifar10 Corrupted dataset \cite{Tarasenko2020StreamingApp}. Finally, STNet robustness to noise was illustrated for adverse weather and lighting conditions. It has been demonstrated that STNet outperforms most of the start-of-the-art models or in the cases of VGG16 delivers very similar accuracy, while been much smaller ($\sim$30x) in terms of number of params \cite{Tarasenko2020V4AS}. 

\section{Novel Architecture: STNet based on down-scaled versions of the state-of-the-art models}

The inspiration for the novel architecture comes from three sources: 1) EfficientNet's successful example of conv net optimization; 2) observation that smaller models are more robust (see section \ref{prior}); 3) STNet's robustness to various types of noise and distortions.

\begin{figure}[t]
\begin{center}
\includegraphics[width=0.8\linewidth]{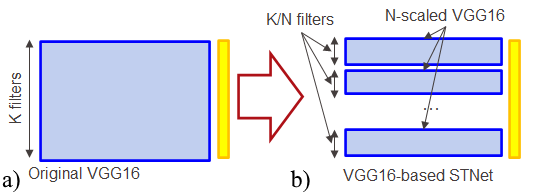}\end{center}
\caption{Down-scaled VGG16-based STNet.}
\label{fig:nscaled}
\end{figure}

The idea to use down-scaled version of VGG16 model as a stream in STNet has been first proposed in \cite{Tarasenko2020StreamingApp}. It has been illustrated that STNet, which used down-scaled VGG16-based streams, has outperformed original VGG16 network in the case of low light image classification. Schema of down-scaled VGG16-based STNet is presented in Fig. \ref{fig:nscaled}. 

In this study, we generalized approach by using down-scaled version of VGG16\cite{Simonyan2014VeryDC}, ResNet\cite{He2015DeepRL} and MobileNetV2 \cite{Howard2017MobileNetsEC}. We refer to these models as to base models. We call an approach of constructing STNet architectures based on down-scaled version of some networks to be a $sparsification$ of such networks.

While MobileNetV2 can be naturally down-scaled using $\alpha$-multiplier, we have designed specialized scalable versions for VGG16 and ResNet50 models. 

In general, we do not require that number of filters in each conv layer of the original network equals total number of filters in corresponding conv layers in STNet architecture.

Next, throughout the paper we name STNet architecture using the following template: STNet$\{$num of streams$\}$$\_\{$scale$\}\_\{$base network name$\}$, where \textit{num of streams} is a number of streams in STNet, 
\textit{scale} is a scale factor, which number of filters in each conv layer of an original model is divided by, scale factor can be any real number, \textit{base network name} is a name of the base network. For example, STNet name STNet(5)$\_$5$\_$ResNet50 describes 5-stream STNet based on 5x down-scaled ResNet50. In the case of MobileNetV2, $scale$ equals $1/\alpha$.

For all the STNet architectures in this paper, we use classifier of the following structure: 1) the outputs of the streams are flatten and concatenated together; 2) result of concatenation is fed into fully connected layer of 400 units with ReLU activation; 3) then batch normalization layer followed by ReLU activation layer are applied; 4) output of ReLU activation layer is fed into the softmax classifier.

\section{Experiments}
\subsection{Model choices}
We have selected VGG16, ResNet50 and MobileNetV2 for experiments. VGG16 and ResNet50 are convolutional nets based on classic convolution layers, while MobileNetV2 uses the most recent advanced technique of depthwise separable convolution. The down-scaled versions of these state-of-the-art networks are used as parallel streams.

\subsection{Objective}
The objective is to find STNet architectures based on VGG16, ResNet50 and MobileNetV2, which outperform or deliver similar accuracy level, while being faster in terms of FLOPs. 


\subsection{Datasets and Protocol}

Cifar10 and Cifar10 Corrupted are used in experiments. Cifar10 is used as training set. We test both base networks and corresponding derived STNets for each type of distortion in Cifar10 Corrupted dataset. For each experiment, we randomly select 50,000 of Cifar10 images. Cifar10 Corrupted also contains 50,000 for each type of distortions. 

We perform two types of test with and without data augmentation. Tests without data augmentation imply that both base model and derived STNet model are trained using Cifar10 original images and tested using corresponding Cifar10 Corrupted samples. Tests with data augmentation imply that 50$\%$ of randomly selected distorted images are added to original training set. The networks are then tested using 50$\%$ of remaining distorted images. We do not use any type of image preprocessing.

The model search is done using tests without data augmentation. The data augmentation tests are performed using selected STNet models.

Throughout the experiments we only use 5-stream STNets. To find the STNet model, which outperforms the base model in terms of accuracy of distorted images classification, we test STNet architectures with the same base model with different scaling factor. We start our tests with STNets, which in total have the same number of filters in each convolutional layer with based networks' convolution layer. For example, STNet5$\_$5$\_$ResNet50 has 5 streams and only 20$\%$ filters of ResNet50 are contained in each stream. The number of filters in the first convolutional layer of ResNet50 equals the total number of filters in the first convolutional layers in all 5 streams. In general, we first test STNet5$\_$5$\_$VGG16, STNet5$\_$5$\_$ResNet50 and STNet5$\_$5$\_$MobileNetV2.

If the first model does not outperform base model, we reduce scale factor and continue tests. We put original constrain that STNet models should be faster in terms of FLOPs than corresponding base models. If we find STNet model which outperforms the base model, we stop tests with such base model. 

\begin{table}
\caption{Base models and corresponding STNet architectures}
\label{models}
\centering
\begin{tabular}{ll}
\hline

Base model & STNet Architecture \\
VGG16 & STNet5$\_$1.5$\_$VGG16 \\
ResNet50 & STNet5$\_$5$\_$ResNet50 \\
MobileNetV2 & STNet5$\_$2.5$\_$MobileNetV2 \\

\hline
\end{tabular}
\end{table}

\begin{table*}
\caption{FLOPs}
\label{model_specs}
\centering
\begin{tabular}{lll}
\hline

Name & FLOPs & Num of Params \\
\hline
(1) STNet5$\_$1.5$\_$VGG16 & 65,829,422 (0.08995) & 32,929,360 (0.979) \\
(2) VGG16 & 731,885,004 & 33,638,218 \\
(3) STNet5$\_$5$\_$ResNet50 & 11,129,542 (0.05436) & 5,593,490 (0.237) \\
(4) ResNet50 & 204,737,884 & 23,608,202 \\
(5) STNet5$\_$2.5$\_$MobileNetV2 & 7,035,742 (0.42068) & 5,093,530 (2.243) \\
(6) MobileNetV2 & 16,724,697 & 2,270,794 \\
\hline
\end{tabular}
\end{table*}


\subsection{Results}

We have obtained three STNet models which outperform or deliver similar performance to the base model. 
Our first test of STNet5$\_$5$\_$ResNet50 has confirmed that this model satisfies requirements of the protocol.

For remaining two models, we have tested various scaling factor value before we could find an appropriate model. In the case of VGG16 network, we have tested STNets with scaling factor of value 5, 4, 3, 2 and 1.5. On the later one was successful resulting in STNet5$\_$1.5$\_$VGG16 architecture. In the case of MobileNetV2, we have tested STNets with scaling factor of value 5 ($\alpha = 0.2$), 3 ($\alpha = 0.33$) and 2.5 ($\alpha = 0.4$). On the later one was successful resulting in STNet5$\_$2.5$\_$MobileNetV2 architecture. 

The base model and corresponding STNet architectures are summarized in Table\ref{models}. The comparison of the number of parameters and FLOPs is presented in Table \ref{model_specs}. In Table \ref{model_specs}, values presented in brackets are ratios between specs of STNet architecture and the ones of corresponding base networks.

One can infer that all the models are much faster in terms of FLOPs. STNet5$\_$1.5$\_$VGG16 and STNet5$\_$5$\_$ResNet50 are also smaller in terms of number of parameters than their correspoding base networks: 0.979x and 0.237x for VGG16 and ResNet50, respectively. On the other hand, STNet5$\_$2.5$\_$MobileNetV2 is nearly twice bigger in terms of parameters for MobileNetV2.

Noise type-wise accuracies for each model are presented in Tables \ref{acc_no_aug} and \ref{acc_aug} for no data augmentation tests and data augmentation tests, respectively. 

\begin{table*}
\caption{Noise Types and Corresponding Accuracies for Tests without Augmentation}
\label{acc_no_aug}
\centering
\begin{tabular}{lclclclclclcl}
\hline

& \multicolumn{6}{c}{Accuracy} \\
\cline{2-7}
Noise & VGG16 & STNet5$\_$ & ResNet50 & STNet5$\_$ & Mobile & STNet5$\_$2.5$\_$ \\ 
Type & & 1.5$\_$VGG16 & & 5$\_$ResNet50 & NetV2 & MobileNetV2 \\
\hline
brightness & 0.771 & 0.746 & 0.498 & 0.595 & 0.379 & 0.397 \\ 
contrast & 0.527 & 0.526 & 0.313 & 0.416 & 0.267 & 0.270 \\ 
defocus blur & 0.725 & 0.698 & 0.524 & 0.601 & 0.416 & 0.426 \\ 
elastic & 0.715 & 0.688 & 0.508 & 0.562 & 0.397 & 0.405 \\ 
transformation & & & & & & \\ 
fog & 0.653 & 0.656 & 0.371 & 0.480 & 0.291 & 0.310 \\ 
frost & 0.723 & 0.689 & 0.437 & 0.543 & 0.325 & 0.345 \\ 
gaussian blur & 0.692 & 0.704 & 0.515 & 0.602 & 0.402 & 0.419 \\ 
gaussian noise & 0.727 & 0.683 & 0.480 & 0.555 & 0.348 & 0.335 \\ 
glass blur & 0.676 & 0.645 & 0.521 & 0.583 & 0.385 & 0.412 \\ 
impulse noise & 0.679 & 0.638 & 0.410 & 0.495 & 0.300 & 0.305 \\ 
jpeg  & 0.770 & 0.747 & 0.546 & 0.623 & 0.417 & 0.436 \\ 
compression & & & & & & \\
motion blur & 0.673 & 0.660 & 0.497 & 0.560 & 0.388 & 0.426 \\ 
pixelate & 0.763 & 0.746 & 0.538 & 0.620 & 0.406 & 0.426 \\ 
saturate & 0.738 & 0.728 & 0.479 & 0.557 & 0.355 & 0.388 \\ 
shot noise & 0.749 & 0.696 & 0.506 & 0.564 & 0.360 & 0.374 \\ 
snow & 0.713 & 0.697 & 0.474 & 0.573 & 0.355 & 0.363 \\ 
spatter & 0.728 & 0.708 & 0.509 & 0.581 & 0.375 & 0.405 \\ 
speckle noise & 0.743 & 0.707 & 0.500 & 0.555 & 0.366 & 0.381 \\ 
zoom blur & 0.699 & 0.669 & 0.503 & 0.579 & 0.374 & 0.404 \\ 
\hline
\end{tabular}
\end{table*}

\textbf{Tests without data augmentation}. It can be concluded from the Table \ref{acc_no_aug} that in most of the case all STNet models deliver similar performance or outperform the base networks. Especially high improvement is observed in the case of STNet5$\_$5$\_$ResNet50. Therefore, we can conclude that faster in terms of FLOPs STNet architectures deliver similar or higher accuracy to classify distorted images. 

Among all the networks, VGG16 and corresponding STNet5$\_$1.5$\_$VGG16 deliver the highest accuracy for all types of noise. STNet5$\_$1.5$\_$VGG16 computations take only 8.9$\%$ of FLOPs than the ones for VGG16 model. Furthermore, STNet5$\_$1.5$\_$VGG16 constitutes 97.9$\%$ of VGG16's size in terms of number of params.

\textbf{Tests with data augmentation}. According to the Table \ref{acc_aug}, in the case of data augmentation, VGG16- and ResNet50-based STNet architectures essentially outperform base models for all types of distortions. In the case of MobileNetV2, STNet architecture deliver at least 5$\%$ accuracy decreased compared with MobileNetV2 in the case of impulse noise and shot noise.

After applying data augmentation, performance of both base models and corresponding derived STNet architectures has essentially improved. To measure the effect of augmentation depending on the type of a model, we compute a difference in performance after and before augmentation for each model for each distortion type. We call this difference \textit{an augmentation boost}. Using the augmentation boost, we can investigate how augmentation effected the performance of the model. 

\begin{figure}[t]
\begin{center}
\includegraphics[width=0.9\linewidth]{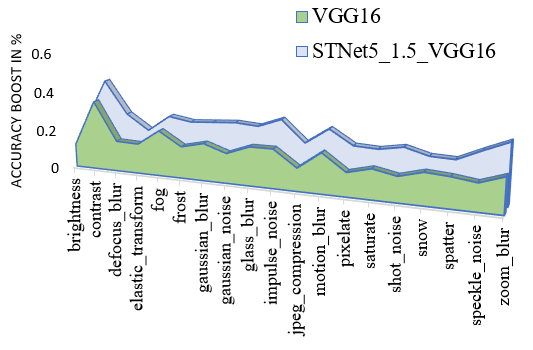}\end{center}
\caption{Boost in accuracy after data augmentation: VGG16 model vs. STNet5$\_$1.5$\_$VGG16}
\label{fig:acc_boost_vgg}
\end{figure}

\begin{figure}[t]
\begin{center}
\includegraphics[width=0.9\linewidth]{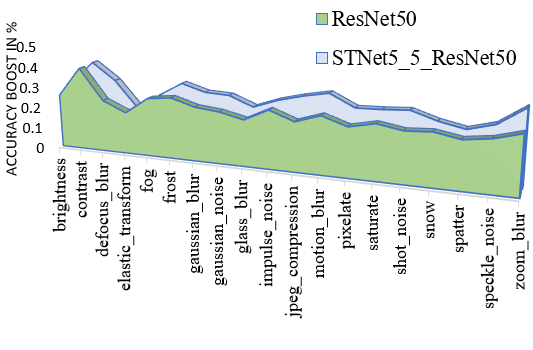}\end{center}
\caption{Boost in accuracy after data augmentation: ResNet50 vs. STNet5$\_$5$\_$ResNet50}
\label{fig:acc_boost_res}
\end{figure}

\begin{figure}[t]
\begin{center}
\includegraphics[width=0.9\linewidth]{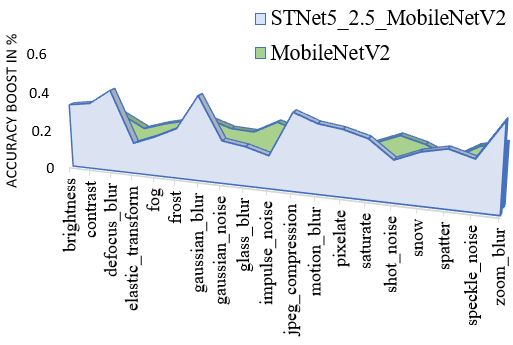}\end{center}
\caption{Boost in accuracy after data augmentation: MobileNetV2 vs. STNet5$\_$2.5$\_$MobileNetV2}
\label{fig:acc_boost_mob}
\end{figure}

In order to compare augmentation boost for each pair of base model and corresponding STNet architecture, we illustrate data for both models in one chart. Figs. \ref{fig:acc_boost_vgg}, \ref{fig:acc_boost_res} and \ref{fig:acc_boost_mob} compare VGG16 model vs. STNet5$\_$1.5$\_$VGG16, ResNet50 vs. STNet5$\_$5$\_$ResNet50 and MobileNetV2 vs. STNet5$\_$2.5$\_$MobileNetV2, respectively.

From \ref{fig:acc_boost_vgg} and \ref{fig:acc_boost_res}, one can infer that in both cases STNet architecture outperforms in accuracy the corresponding base model for all types of noise. However, in the case of MobileNetV2 vs. STNet5$\_$2.5$\_$MobileNetV2 pair, base model outperforms corresponding STNet architecture in augmentation boost for elastic transform, fog, frost, gaussian noise and blur, glass blur, impulse noise, shot noise and speckle noise. This illustrates limitations of the STNet, when MobileNetV2 is used as a base model.

Finally, we found that STNet5$\_$1.5$\_$VGG16 is the best model with pronounced supremacy in accuracy over all other networks. STNet5$\_$1.5$\_$VGG16 also exhibits the highest data augmentation boost among all the models across all the noise types.

\begin{table*}
\caption{Noise Types and Corresponding Accuracies for Tests with Augmentation}
\label{acc_aug}
\centering
\begin{tabular}{lclclclclclcl}
\hline

& \multicolumn{6}{c}{Accuracy} \\
\cline{2-7}
Noise & VGG16 & STNet5$\_$ & ResNet50 & STNet5$\_$ & Mobile & STNet5$\_$2.5$\_$ \\ 
Type & & 1.5$\_$VGG16 & & 5$\_$ResNet50 & NetV2 & MobileNetV2 \\
\hline
brightness & 0.890 & 0.966 & 0.753 & 0.891 & 0.656 & 0.726 \\ 
contrast & 0.881 & 0.965 & 0.708 & 0.819 & 0.599 & 0.616 \\ 
defocus blur & 0.883 & 0.972 & 0.772 & 0.924 & 0.682 & 0.849 \\ 
elastic & 0.869 & 0.885 & 0.706 & 0.755 & 0.604 & 0.563 \\ 
transformation & & & & & & \\ 
fog & 0.887 & 0.938 & 0.648 & 0.749 & 0.534 & 0.513 \\ 
frost & 0.887 & 0.952 & 0.728 & 0.877 & 0.594 & 0.601 \\ 
gaussian blur & 0.884 & 0.976 & 0.771 & 0.903 & 0.690 & 0.852 \\ 
gaussian noise & 0.880 & 0.963 & 0.725 & 0.851 & 0.598 & 0.552 \\ 
glass blur & 0.875 & 0.918 & 0.739 & 0.833 & 0.629 & 0.612 \\ 
impulse noise & 0.879 & 0.927 & 0.684 & 0.788 & 0.604 & 0.472 \\ 
jpeg  & 0.888 & 0.958 & 0.779 & 0.945 & 0.686 & 0.827 \\ 
compression & & & & & & \\
motion blur & 0.878 & 0.953 & 0.768 & 0.905 & 0.681 & 0.773 \\ 
pixelate & 0.884 & 0.967 & 0.770 & 0.908 & 0.683 & 0.755 \\ 
saturate & 0.890 & 0.944 & 0.737 & 0.848 & 0.600 & 0.685 \\ 
shot noise & 0.878 & 0.934 & 0.743 & 0.863 & 0.653 & 0.581 \\ 
snow & 0.876 & 0.902 & 0.718 & 0.834 & 0.616 & 0.620 \\ 
spatter & 0.882 & 0.910 & 0.733 & 0.818 & 0.580 & 0.690 \\ 
speckle noise & 0.882 & 0.964 & 0.739 & 0.824 & 0.643 & 0.630 \\ 
zoom blur & 0.880 & 0.979 & 0.776 & 0.931 & 0.677 & 0.840 \\ 
\hline
\end{tabular}
\end{table*}



\section{Discussion and conclusion}

Our results indicate that the network with the highest accuracy is STNet5$\_$1.5$\_$VGG16 delivering start-of-the-art results. This network is essentially smaller and faster than original VGG16. For example, STNet5$\_$1.5$\_$VGG16 computations take only 8.9$\%$ of VGG16 computations in FLOPs. The possible reason is the original VGG16 has two fully connected layers of size 4094 each. This constitutes the major amount of computations. 

We have also shown that STNet architectures can deliver higher or similar accuracy for distorted image classification for under no data augmentation conditions for all three selected base models, i.e., VGG16, ResNet50 and MobileNetV2. On the other hand, for data augmentation test STNets based on VGG16 and ResNet50 models exhibit essential augmentation boost and outperform base models in accuracy for all types of noise. However, for MobileNetV2 in the case of impulse and shot noise, it was not possible to achieve accuracy improvement over the base model.

In all the cases, STNet architecture took much less computations in terms of FLOPs. STNets based on VGG16 and ResNet50 are also smaller in terms of number of params. However, in the case of MobileNetV2, selected model was nearly twice bigger.

The worst performance was exhibited by MobileNetV2 for all types of distortions. This suggests that networks based on classic convolution layers like VGG16 and ResNet50 are more robust to noise than MobileNetV2. In general, STNet architectures can essentially increase robustness to noise for all types of noise in the case of VGG16 and ResNet50 as base models. The applications of MobileNetV2-based STNets is limited. 

We speculate here that after training on any dataset, a network becomes fine-tuned to a given dataset. There is always a question how well the trained network generalizes beyond the samples seen during the training. Especially, when network trained on one dataset and then tested on a different dataset, it is important to understand how much a trained network is ``overfitted" to the first dataset. Overall, STNet models demonstrate a high level of generalization to unseen samples in the case of distorted images.

In the paper \cite{Tarasenko2020STNets}, it was illustrated that input-induced sparsity (image intensity slices used as input into different streams) and hard-wired sparsity (parallel weight-decoupled streams) are both necessary for noise robustness to emerge in multi-stream architectures based on a simple conv net. STNet model based on the state-of-the-art models are build upon both input-induced sparsity and hard-wired sparsity. Thus, we have proved that the same principles are correct in the case of state-of-the-art models. Therefore, our experimental results illustrate the STNet is a flexible architecture capable of increasing noise robustness and applicable for a variety of networks. The effective models for can be found using simple search by varying scaling factor value.

Finally, our approach as well as EfficientNet optimization deals with both network specs (width in terms of parallel streams and number of filters) and input specs (intensity slices). In contrast to EfficientNet’s optimization, we focus on FLOPs and noise robustness, which is essentially an accuracy on dataset derived from a different statistical distribution than a training set, rather than accuracy on a test set derived from the same statistical distribution as training set. Therefore, we are looking for models with higher generalization ability for previously unseen samples.

Summarizing, we conclude that STNet is a flexible (applicable for a variety of model) and efficient (in terms of FLOPs and number of params) architecture, which enables to increase robustness to various distortions. Therefore STNet is a promising tool to make original neural network models more robust to noise and other types of distortions.

\end{document}